%% file: main.tex
\documentclass[10pt,twocolumn,letterpaper]{article}

\usepackage{iccv}              


\definecolor{iccvblue}{rgb}{0.21,0.49,0.74}
\usepackage[pagebackref,breaklinks,colorlinks,allcolors=iccvblue]{hyperref}
\usepackage{microtype}
\usepackage{graphicx}
\usepackage{adjustbox}
\usepackage{hhline}
\usepackage{colortbl}
\usepackage{tabularray}
\usepackage{physics}
\usepackage{hyperref}
\usepackage{url}
\usepackage{booktabs}       
\usepackage{amsfonts}       
\usepackage{nicefrac}       
\usepackage{xcolor}         
\usepackage{pifont}
\newcommand{\cmark}{\ding{51}}%
\newcommand{\xmark}{\ding{55}}%
\usepackage{multirow} 
\usepackage{mdframed}
\usepackage{fontawesome}
\usepackage{amsmath}
\usepackage{amssymb}
\usepackage{mathtools}
\usepackage{amsthm}

\def\paperID{4453} 
\def\confName{ICCV}
\def\confYear{2025}

\title{3DRealCar: An In-the-wild RGB-D Car Dataset with 360-degree Views}

\author{
  Xiaobiao Du \textsuperscript{1,2,3}\space \space \space \space
  Yida Wang   \textsuperscript{3}\ \space \space \space \space
  Haiyang Sun \textsuperscript{3}\ \space \space \space \space
 Zhuojie Wu   \textsuperscript{1} \space \space \space \space
 Hongwei Sheng   \textsuperscript{1} \\
   Shuyun Wang \textsuperscript{1}  \space \space \space \space
  Jiaying Ying \textsuperscript{1} \space \space \space \space
  Ming Lu \textsuperscript{5} \space \space \space \space
 Tianqing Zhu \textsuperscript{4} \space \space \space \space
 Kun Zhan \textsuperscript{3} \space \space \space \space
  Xin Yu\textsuperscript{1}\thanks{Corresponding author.}  \\
  \\
  \textsuperscript{1}  The University of Queensland\space \space
    \textsuperscript{2}  University of Technology Sydney \space \space
    \textsuperscript{3}  Li Auto Inc. \\ 
    \textsuperscript{4} City University of Macau  \space \space
  \textsuperscript{5} Peking University \\
}

\begin{document}
\maketitle

\input{sec/abstract}

\input{sec/intro}

\input{sec/related}

\input{sec/dataset}

\input{sec/task}

\input{sec/exp}

\input{sec/conclusion}

{
    \small
    \bibliographystyle{ieeenat_fullname}
    \bibliography{main}
}

\end{document}

%% file: sec/abstract.tex
\begin{figure*}[ht]
    \centering
    \includegraphics[width=0.95 \linewidth]{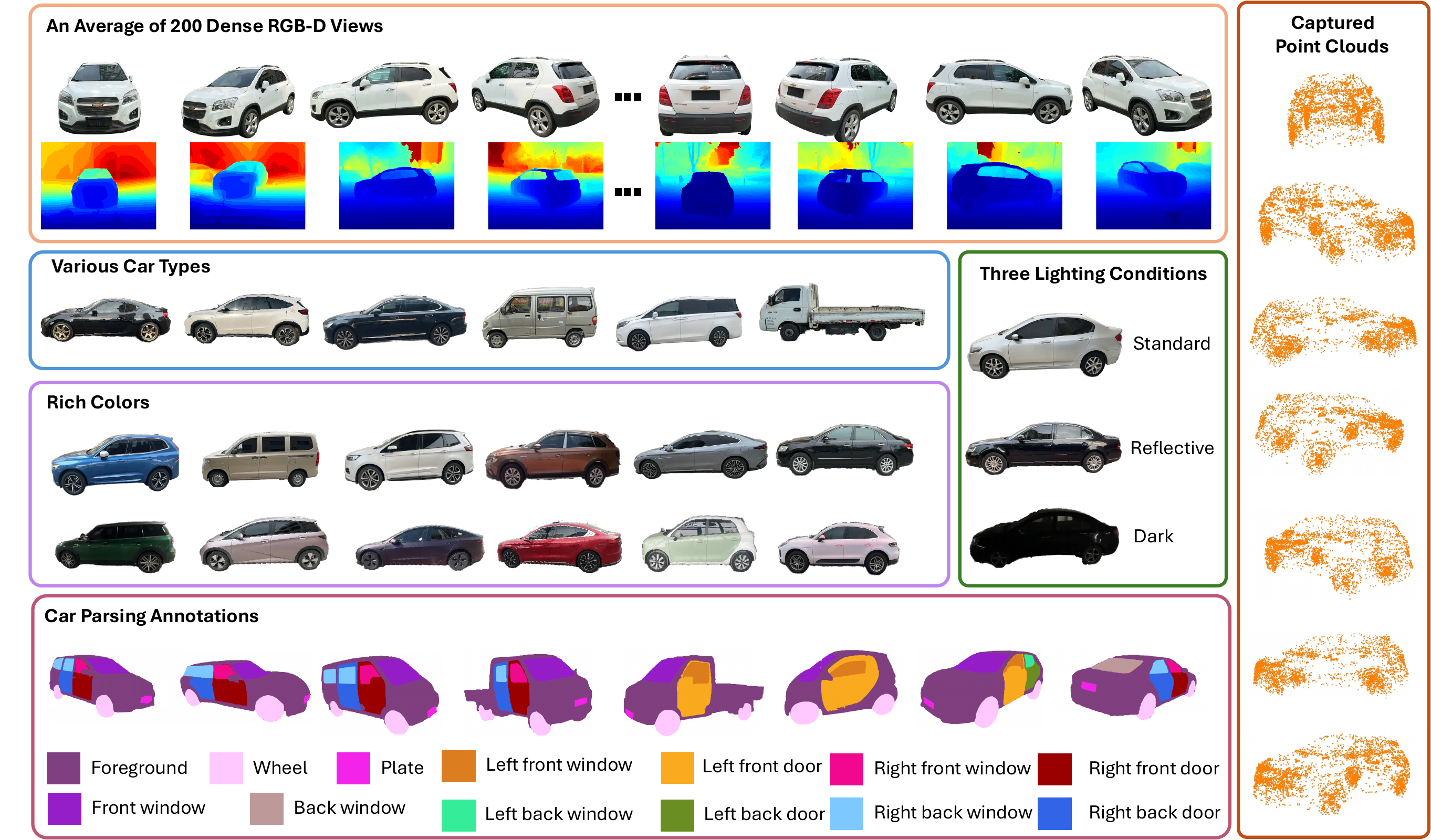} 
    \caption{\textbf{Characteristics of our curated high-quality 3DRealCar dataset.} 3DRealCar contains detailed annotations for various colors, car types, brands, and car parsing maps. 3DRealCar contains three lighting conditions on car surfaces, bringing challenges to existing methods.}
    \label{teaser}
    \vspace{-4mm}
\end{figure*}

\begin{abstract}

3D cars are commonly used in self-driving systems, virtual/augmented reality, and games. However, existing 3D car datasets are either synthetic or low-quality, limiting their applications in practical scenarios and presenting a significant gap toward high-quality real-world 3D car datasets. In this paper, we propose the first large-scale 3D real car dataset, termed 3DRealCar, offering three distinctive features. (1) \textbf{High-Volume}: 2,500 cars are meticulously scanned by smartphones, obtaining car images and point clouds with real-world dimensions; (2) \textbf{High-Quality}: Each car is captured in an average of 200 dense, high-resolution 360-degree RGB-D views, enabling high-fidelity 3D reconstruction; (3) \textbf{High-Diversity}: The dataset contains various cars from over 100 brands, collected under three distinct lighting conditions, including reflective, standard, and dark. Additionally, we offer detailed car parsing maps for each instance to promote research in car parsing tasks. Moreover, we remove background point clouds and standardize the car orientation to a unified axis for the reconstruction only on cars and controllable rendering without background. We benchmark 3D reconstruction results with state-of-the-art methods across different lighting conditions in 3DRealCar. Extensive experiments demonstrate that the standard lighting condition part of 3DRealCar can be used to produce a large number of high-quality 3D cars, improving various 2D and 3D tasks related to cars. Notably, our dataset brings insight into the fact that recent 3D reconstruction methods face challenges in reconstructing high-quality 3D cars under reflective and dark lighting conditions. 
\textcolor{red}{\href{https://xiaobiaodu.github.io/3drealcar/}{Our dataset is here.}}



\end{abstract}

%% file: sec/intro.tex
\input{table/data_compr}

\begin{figure}[t] 
    \centering
    \includegraphics[width=0.48\textwidth]{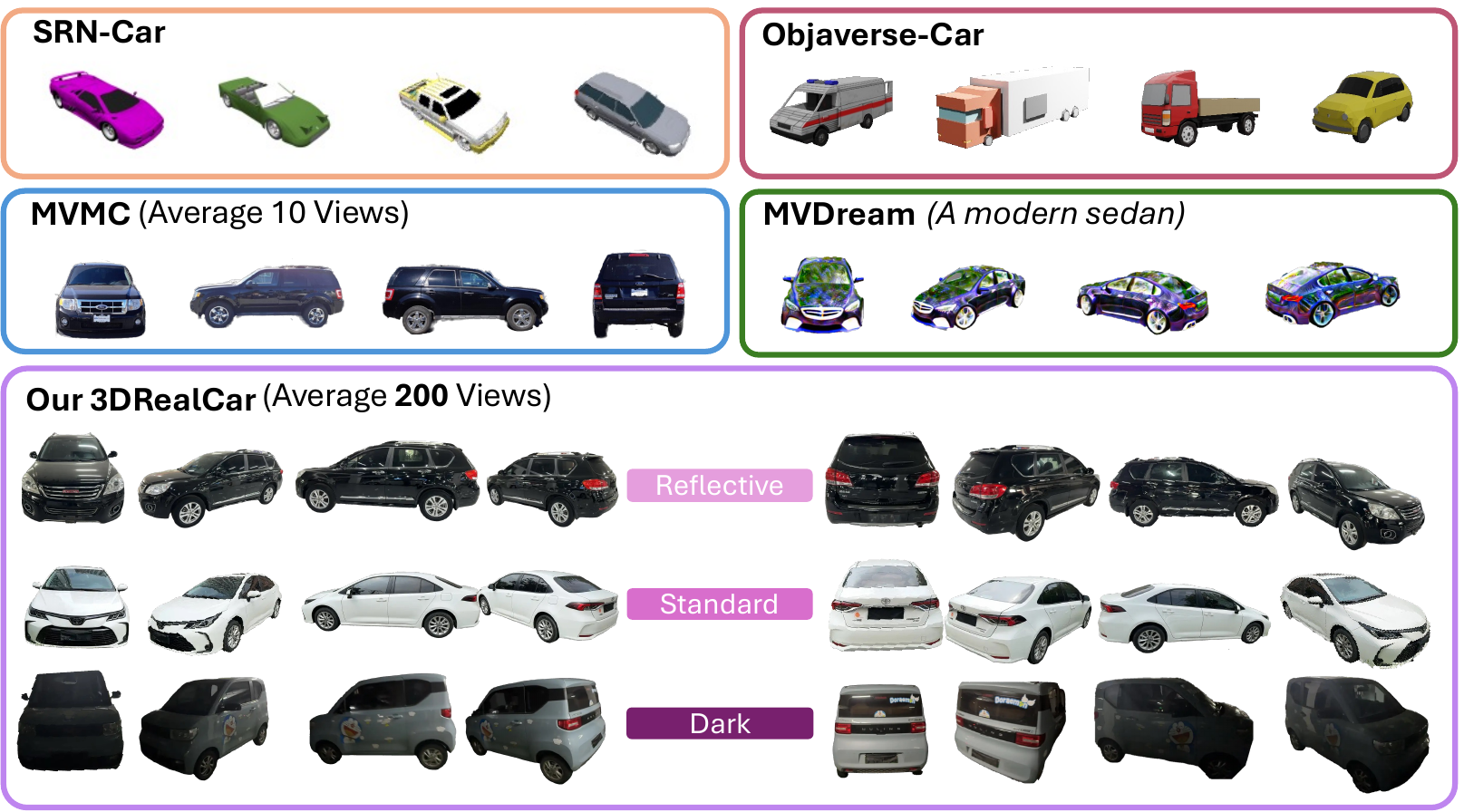} 
    \caption{\textbf{Visual comparisons of 3D car datasets and the results of a 3D generative method.} Our 3DRealCar is captured in real-world scenes and contains more densely captured views. In addition, our dataset has annotations for three different lighting conditions on the car surface. We also compare a recent state-of-the-art text-to-3D model, MVDream~\citep{shi2023mvdream} with a prompt ``\textit{a modern sedan}'', demonstrating its failure to generate high-quality 3D car models.}
    \label{data_comp}
    \vspace{-5mm}
\end{figure}

\section{Introduction}
Cars, as both daily objects and vehicles, are of significant interest to researchers, especially in the field of autonomous driving. 
Autonomous perception systems are typically trained on daily scene datasets that are collected frequently. However, these datasets often exhibit long-tailed distributions, with far fewer instances of corner-case scenarios, like car accidents.
Consequently, this imbalance leads to the autonomous perception system generalizing well in the most frequently occurring scenes. This means that the system is likely to perform poorly in rare situations, posing significant safety risks to drivers. To build a reliable system, it is essential to have a simulator that can simulate photorealistic hazardous scenes. Moreover, high-quality 3D cars are necessary for a realistic simulator.

Recent 3D car reconstruction methods mainly reconstruct cars from self-driving datasets. 
To apply reconstructed cars to real-world scenes, the reconstructed 3D car should be high-quality. 
However, it is very challenging to obtain such high-quality 3D cars for the following reasons: (1) Previous 3D car reconstruction methods produce low-quality 3D cars, primarily because they train on self-driving datasets with low-resolution car images and a limited number of trainable views. (2) Manually crafting a high-quality 3D car model requires specialized artists, which is time-consuming. (3) There is no large-scale 3D real car dataset that can be utilized to produce a bulk of 3D cars.

Moreover, existing 3D car datasets are either synthetic or only contain a few posed images, as shown in Figure \ref{data_comp}. SRN-Car~\citep{chang2015shapenet} and Objaverse-Car~\citep{deitke2023objaverse} collect 3D car computer-aided design (CAD) models from the Internet, but these models are synthetic and contain non-photorealistic texture. Although MVMC~\citep{zhang2021ners} is a real car dataset, it collects only ten views on average for each car. 
On the contrary, our collected 3DRealCar dataset provides an average of 200 dense RGB-D views per car for high-quality 3D car reconstruction.

We also show that the recent state-of-the-art 3D generative method, MVDream~\citep{shi2023mvdream}, as depicted in Figure \ref{data_comp}, fails to generate high-quality cars due to the multi-view inconsistency introduced by generative models.
Thus, the existing 3D generation methods cannot be employed to generate high-quality 3D real car assets.

In this work, we collect a large-scale 3D real car dataset in the wild, termed 3DRealCar, which contains dense high-quality views and rich diversity. During data collection, we employ smartphones with ARKit~\citep{arkit} to scan cars parked on roadsides or parking lots, obtaining posed RGB-D images and point clouds of cars. In particular, we scan around the cars in three loops to obtain dense views. Note that we collect car data with the consent of owners. In Table \ref{t1} and Figure \ref{teaser}, we show our dataset possesses striking characteristics compared with previous 3D car datasets. We capture dense RGB-D images in high resolution, which promotes the reconstruction of high-quality 3D cars. Furthermore, we scan cars under three different lighting conditions, resulting in the surfaces of cars having different lighting effects, such as reflective, standard, and dark, where we denote the standard as the smooth lighting condition without obvious specular highlight. Figure \ref{data_comp} shows some examples of three lighting conditions in our dataset.
Note that the number of instances in our dataset is the largest in existing datasets. 
Therefore, our collected 3DRealCar dataset has a rich diversity in terms of car types, colors, brands, and lighting conditions. 
We also provide car parsing map annotations with thirteen classes for each instance, which enable our dataset to be applied in car component understanding tasks.

To construct a high-quality dataset, we filter out the images that are out of focus, occluded, or blurred.
To facilitate the 3D reconstruction solely on cars, we remove the point clouds of the background. We also adjust the orientation of the car facing along the x-axis before the reconstruction for controllable rendering.
Based on the high-quality posed RGB-D images, point clouds, and multi-grained annotations, we can apply the dataset to various tasks related to cars.
Figure \ref{fig_tasks} shows our dataset supports over 10 tasks, including several popular 2D and 3D tasks to promote the advancement of car-related research.

We leverage existing state-of-the-art methods to benchmark 3D reconstruction and car parsing tasks of our 3DRealCar dataset.
We also conduct extensive experiments to demonstrate that the reflective and dark lighting conditions in our dataset are challenging to existing methods, which brings a new challenge for 3D reconstruction in awful lighting conditions. Furthermore, we demonstrate that our 3DealCar dataset can bring real-car prior and enhance existing 3D generation and downstream methods. 
Overall, the contributions of this work can be summarized below:

$\bullet$  We propose the first large-scale 3D real car dataset, named 3DRealCar, which contains 2,500 car instances and their point clouds with actual sizes in real-world scenes.

$\bullet$  3DRealCar contains RGB-D images and point clouds with detailed annotations, supporting researchers to investigate various tasks in both 2D and 3D.

$\bullet$  We conduct 3D reconstruction and car parsing benchmarks to advance car-related tasks. Notably, we observe that existing methods face challenges under the extreme lighting conditions of 3DRealCar.

$\bullet$ Extensive experiments demonstrate 3DRealCar can enhance real-car prior and improve the performance of existing 3D generation and novel view synthesis methods.

%% file: table/data_compr.tex
\begin{table*}[]
\caption{\textbf{The comparison of existing 3D car datasets.} Our dataset contains unique characteristics compared with existing 3D car datasets. Lighting means the lighting conditions of the surfaces of cars. Point Cloud represents the point clouds with actual sizes in real-world scenes.
}
\label{t1}
\centering
\scriptsize
\def\arraystretch{1.2}

\setlength{\tabcolsep}{0.6em}{
\begin{tabular}{lccccccccc}
\toprule \toprule
\textbf{Dataset}          & \textbf{Instances} & \textbf{Type} & \textbf{Views}     & \textbf{Resolution} & \textbf{Brand} & \textbf{Lighting} & \multicolumn{1}{l}{\textbf{Car Parsing}} & \textbf{Depth} & \textbf{Point Cloud} \\ \midrule 
SRN-Car                   & 2151               & Synthetic     & 250                & 128$\times$128                   & \xmark              & \xmark               & \xmark                               & \xmark          & \xmark             \\
Objaverse-car             & 511                & Synthetic     & -                  & -                   & \xmark               & \xmark               & \xmark                                & \xmark          & \xmark           \\
MVMC                      & 576                & Real          & $\sim$10           & 600$\times$450             & $\sim$40              & \xmark              & \xmark                                 & \xmark      & \xmark             \\ \midrule
\textbf{3DRealCar (Ours)} & \textbf{2500}      & \textbf{Real} & \textbf{$\sim$200} & \textbf{1920$\times$1440}  & \textbf{100+}  & \textbf{3}     & \textbf{13 }                                & \cmark          & \cmark            \\ \bottomrule
\end{tabular}
}
\vspace{-3mm}
\end{table*}

%% file: sec/related.tex
\begin{figure}[t] 
    \centering
    \includegraphics[width=0.48\textwidth]{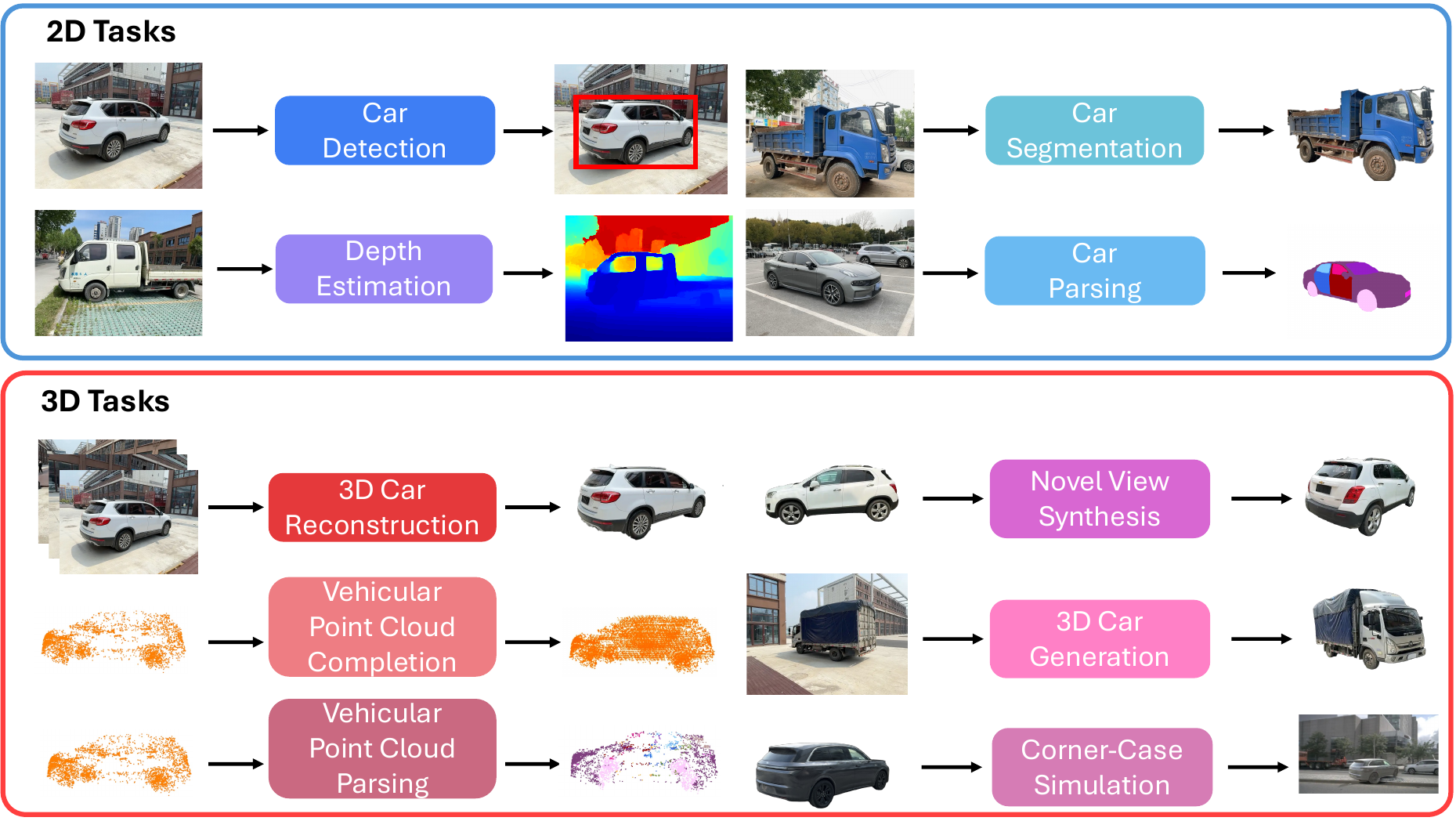} 
    \caption{\textbf{The applicable tasks of our dataset.} Our proposed 3DRealCar dataset containing RGB-D images, point clouds, and rich annotations, can be applied to various popular 2D and 3D tasks to support the construction of safe and reliable self-driving systems. }
    \label{fig_tasks}
    \vspace{-5mm}
\end{figure}

\section{Related Work}

\noindent\textbf{3D Car Datasets.} There are several well-known large-scale autonomous driving datasets so far, such as Nuscenes, KITTI, Waymo, Pandaset~\citep{xiao2021pandaset}, ApolloScape~\citep{huang2018apolloscape}, and Cityscape. These datasets are captured by multi-view cameras and lidars mounted on ego cars. Various works \citep{wang2023cadsim, garcia2001neusim, du2024dreamcar} attempt to reconstruct 3D cars in these datasets. However, these methods fall short of reconstructing high-quality 3D cars due to the lack of sufficient and dense training views. 
SRN-Car \citep{chang2015shapenet} and Objaverse~\citep{deitke2023objaverse} collect
3D car models from existing repositories and Internet sources.
However, these datasets only contain synthetic cars, which cannot produce realistic textures and geometry. MVMC~\citep{zhang2021ners} is collected from car advertising websites, which contain a series of car images, especially multi-view images of each car. However, the views of images per car in MVMC are unposed and sparse, which is adverse to reconstructing high-quality 3D car models.
In this paper, we collect a high-quality 3D real car dataset to fill the above gaps.

\noindent\textbf{3D Reconstruction with Neural Field.} 3D reconstruction aims to create a 3D structure digital representation of an object or a scene from its multi-view images, which is a long-standing task in computer vision.
One of the most representative works in 3D reconstruction is Neural Radiance Fields (NeRFs)~\citep{mildenhall2021nerf}, which demonstrates promising performance for novel view synthesis.
Afterward, this method inspires a new wave of 3D reconstruction methods using the volume rendering method, with subsequent works focusing on improving its quality~\citep{ wang2023f2nerf}, efficiency~\citep{ chen2022tensorf}, applying artistic effects, and generalizing to unseen scenes. Particularly, Kilonerf~\citep{reiser2021kilonerf} accelerates the training process of NeRF by dividing a large MLP into thousands of tiny MLPs. Furthermore, Mip-NeRF~\citep{barron2021mip} proposes a conical frustum rather than a single ray to ameliorate aliasing. Mip-NeRF 360~\citep{Barron2022MipNeRF3U} further improves the application scenes of NeRF to the unbounded scenes. Although these NeRF-based methods demonstrate powerful performance on various datasets, the training time always requires several hours even one day more. Instant-NGP~\citep{muller2022instant} uses a multi-resolution hash encoding method, which reduces the training time by a large margin. 
3DGS~\citep{kerbl20233d} proposes a new representation based on 3D Gaussian Splatting, which reaches real-time rendering for objects or unbounded scenes. 
2DGS~\citep{2dgs} proposes a perspective-accurate 2D splatting process that leverages ray-splat intersection and rasterization to further enhance the quality of the reconstructions. 
Scaffold-GS~\citep{lu2023scaffold} proposes an anchor growing and pruning strategy to accelerate the scene coverage.
MVGS~\cite{du2024mvgs} firstly proposes the multi-view training strategy to optimize 3DGS in a more comprehensive way for holistic supervision.
However, there is not yet a large-scale 3D real car dataset so far. Therefore, we present a 3D real car dataset, named 3DRealCar in this work.

\noindent\textbf{3D Generation with Diffusion Prior.}
Some current works~\citep{jun2023shape,nichol2022pointe} leverage a 3D diffusion model to learn the representation of 3D structure. However, these methods lack generalization ability due to the scarcity of 3D data. 
To facilitate 3D generation without direct supervision of  3D data, image or multi-view diffusion models are often used to guide the 3D creation process.
Notable approaches like DreamFusion~\citep{poole2022dreamfusion} and subsequent works~\citep{metzer2023latent} use an existing image diffusion model as a scoring function, applying Score Distillation Sampling SDS loss to generate 3D objects from textual descriptions. 
These methods, however, suffer from issues such as the Janus problem~\citep{poole2022dreamfusion} and overly saturated textures. 
Inspired by Zero123~\citep{zero123}, several recent works~\citep{stable-zero123,shi2023zero123++} refine image or video diffusion models to better guide the 3D generation by producing more reliable multi-view images. However, these generative methods fail to generate high-quality cars without the prior of real cars.

%% file: sec/dataset.tex
\begin{figure*}[t!] 
    \centering
    \includegraphics[width=0.95\textwidth]{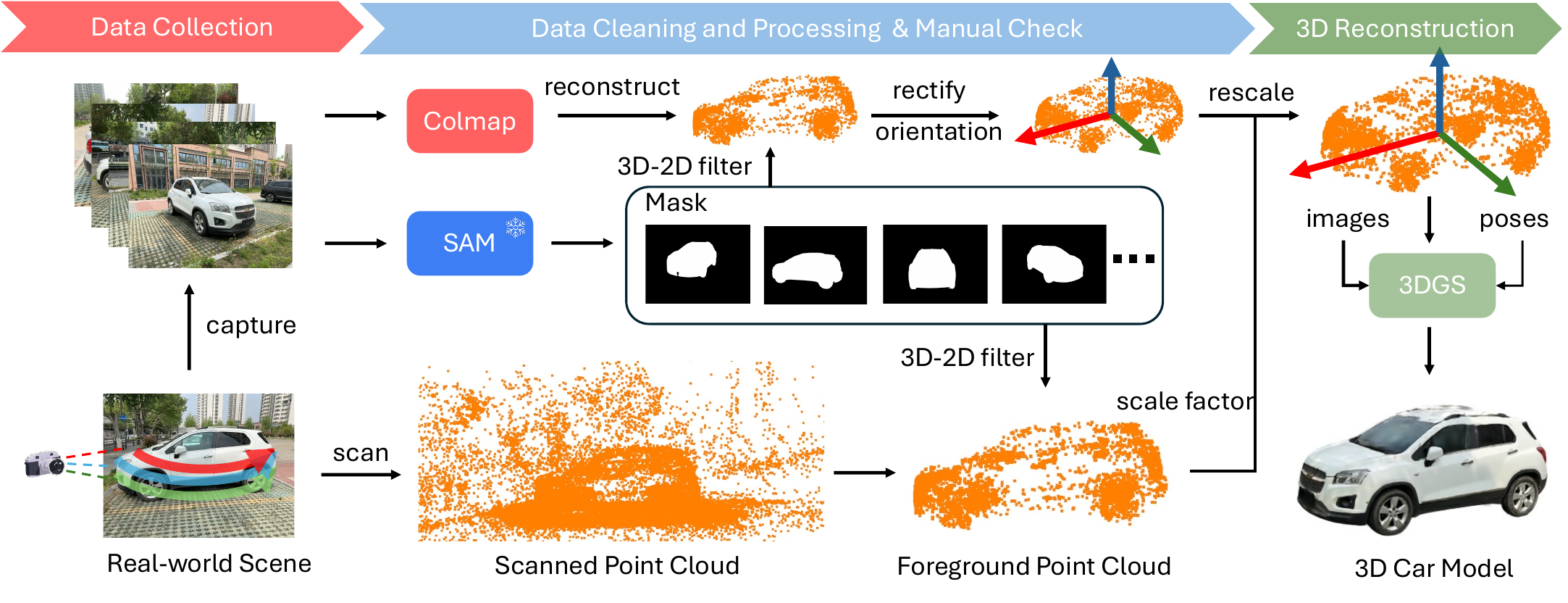} 
    \caption{\textbf{Illustration of our data collection and preprocessing.} We first circle a car three times while scanning the car
with a smartphone for the attainment of RGB-D images and its point clouds. Then we use Colmap~\citep{sfm2016} and SAM~\citep{sam} to obtain poses and remove the background point clouds. Finally, we use the 3DGS~\citep{kerbl2023gaussiansplatting} trained on the processed data to obtain 3D car model. }
    \label{pipeline}
    \vspace{-5mm}
\end{figure*}

\section{Proposed 3DRealCar Dataset}


\subsection{Data Collection and Annotation}
As shown in Figure \ref{pipeline}, our dataset is collected using smartphones, specifically iPhone 14 models, adopting ARKit APIs~\citep{arkit} to scan cars for their point clouds and RGB-D images. The data collection process is conducted under three distinct lighting conditions, such as standard, reflective, and dark. These lighting conditions represent the lighting states of vehicle surfaces. It is important to note that all data collection is performed with the consent of owners. During the scanning process, the car should be stationary while we meticulously circle the car three times to capture as many views as possible. For each loop, we adjust the height of the smartphone to obtain images from different angles. Furthermore, we try our best to make sure captured images contain the entire car body without truncation. To preserve the privacy of owners, we make license plates and other private information obfuscated. To construct a high-quality dataset, we filter out some instances with blurred, out-of-focus, and occluded images. We also provide detailed annotations for car brands, types, and colors. Particularly, we provide the car parsing maps for each car with thirteen classes in our dataset as shown in Figure \ref{teaser} for the advancement of car component understanding tasks.

\subsection{Data Preprocessing}

\noindent\textbf{ Background Removal.} Since we only reconstruct cars for the 3D car reconstruction task, the background should be removed. 
Recent Segment Anything Model (SAM)~\citep{sam} demonstrates powerful context recognition and segmentation performance. However, SAM needs a bounding box, text, or point as a driving factor for accurate segmentation. Therefore, we employ Grounding DINO \citep{liu2023grounding} as a text-driven detector with a detection prompt with ``car'' for the attainment of car bounding boxes. With these bounding boxes, we use SAM to obtain the masks from captured images.
The point cloud initialization is demonstrated useful for the convergence of 3D Gaussian Splatting \citep{kerbl2023gaussiansplatting}.
Except for the removal of the background in 2D images, we still need to remove the background point clouds. Therefore, we first project the 3D point clouds into 2D space with camera parameters. Then, we can eliminate background point clouds with masks and save them for further processing.

\noindent\textbf{Orientation Rectification.} As shown in Figure \ref{pipeline}, we utilize Colmap~\citep{sfm2016} to reconstruct more dense point clouds and obtain accurate camera poses and intrinsics because we find that the estimated poses by the smartphone are not accurate. However, after the removal of the background point clouds, we find that the car orientation of the point cloud is random, which leads to the subsequent render task being uncontrollable. Given camera poses $P = \{ p_i \}_1^\mathcal{N}$, where  $\mathcal{N}$ is the number of poses, we use Principal Component Analysis (PCA)~\citep{abdi2010principal} to obtain a PCA component $\mathcal{T}\in \mathbb{R}^{3\times3}$.
The PCA component is the principal axis of the data in 3D space, which represents rotation angles to each axis. Therefore, we leverage it to rectify the postures of cars parallel to the x-axis. However, this process cannot guarantee cars facing along the x-axis. Therefore, in some failure cases, we manually interfere and adjust the orientation along the x-axis. With the fixed car orientation, we can control rendered poses for the subsequent tasks.

\noindent\textbf{Point Cloud Rescaling.} The size of the point clouds reconstructed by Colmap~\citep{sfm2016} does not match the real-world size, which inhibits the reconstruction of a practically sized 3D car. To address this, we calculate the bounding box of the scanned foreground point clouds to obtain its actual size in the real-world scene. Then, we rescale the rectified point clouds into the real size. In addition to the rescaling of the point clouds, we also need to adjust the camera poses. We rescale translations of camera poses using a scale factor calculated by the ratio of scanned point cloud size and Colmap point cloud size. After these rescaling processes, we use rescaled point clouds to reconstruct a 3D car model through recent state-of-the-art methods, like 3DGS~\citep{kerbl2023gaussiansplatting}.


\subsection{Data Statistics}
In our 3DRealCar, we provide detailed annotations for researchers to leverage our dataset for different tasks. During the data annotating, we discard the data with the number of views less than fifty. As we can observe in Figure \ref{teaser} and \ref{data_comp}, we collect our dataset under real-world scenes and meticulously scan dense views. Therefore, cars in our dataset possess dense views and realistic texture, which is necessary for the application in a real-world setting. 

\begin{figure*}[t] 

    \centering

    \includegraphics[width=0.95\textwidth]{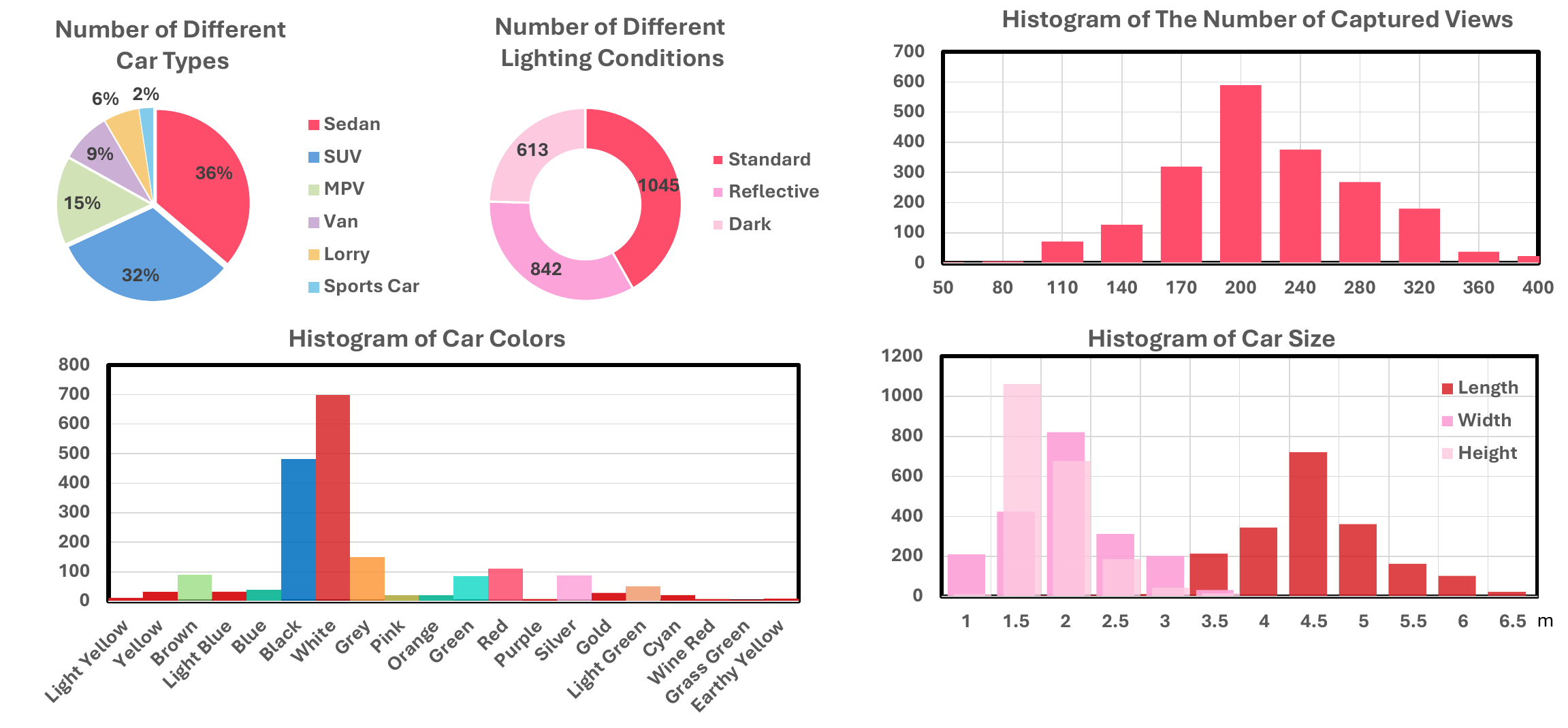} 
    \caption{\textbf{The distributions of our 3DRealCar dataset.} 
    We show distributions of car types, lighting conditions, captured views, car colors, and car size. We try our best to capture cars with various colors and types for the diversity of our dataset. 
    }
    \label{statistics}
    \vspace{-5mm}
\end{figure*}

As shown in Figure \ref{statistics}, we conduct detailed statistical analyses to show the features of our dataset. Our dataset mainly contains six different car types, such as Sedan, SUV, MPV, Van, Lorry, and Sports Car. Among them, sedans and SUVs are common in real life, so their volume dominates in our dataset. We also count the number of different lighting conditions on cars. The standard condition means the car is well-lit and without strong highlights. The reflective condition means the car has strong specular highlights. Glossy materials bring huge challenges to recent 3D reconstruction methods. The dark condition means the car is captured in an underground parking so not well-lit. To promote high-quality reconstruction, we save the captured images in high resolution (1920$\times$1440) and also capture as many views as possible. The number of captured images per car is an average of 200. The number of views ranges from 50 to 400. To enrich the diversity of our dataset, we try our best to collect as many different colors as possible. Therefore, our dataset contains more than twenty colors, but the white and black colors still take up most of our dataset. In addition, we also show the distribution of car size, in terms of their length, width, and height. We obtain their sizes by computing the bounding boxes of the scanned point clouds. Thanks to different car types, the sizes of cars are also diverse.

%% file: sec/task.tex
\section{Overview of 3DRealCar Tasks}
\label{sec_task}
\subsection{2D tasks}

\noindent\textbf{Corner-case scene 2D Detection}~\citep{yolov8}: Given images $I = \{ I_i  \}_1^\mathcal{N} $, this task aims to detect vehicles as accurately as possible. However, in some corner cases, like car accidents, detectors sometimes fail to detect target vehicles since this kind of scene is rare or not in the training set. Therefore, this task has crucial significance in building a reliable self-driving system, especially for accident scenarios.

\noindent\textbf{2D Car Parsing}~\citep{wang2023internimage, ddrnet, xie2021segformer, liu2024vmamba}: Given a serial of images $I = \{ I_i  \}_1^\mathcal{N} $, this task aims to segment car parsing maps $S = \{ S_i  \}_1^\mathcal{N} $. With annotated parsing maps and images, we can train a model to understand and segment each component of cars. This task can assist self-driving systems with more precise recognition.

\subsection{3D Tasks}

\noindent\textbf{Neural Field-based Novel View Synthesis}~\citep{instant-ngp, kerbl2023gaussiansplatting,2dgs}: Given a serial of images $I = \{ I_i  \}_1^\mathcal{N} $ and matched poses $P = \{ p_i \}_1^\mathcal{N}$, where  $\mathcal{N}$ is the number of images and poses, the task of Neural Field-based Novel View Synthesis aims to reconstruct Neural Field model of a object or a scene. The reconstructed model is usually used to render 2D images with different views for the evaluation of the performance of novel view synthesis.

\noindent\textbf{Diffusion-based Novel View Synthesis}~\citep{zero123, liu2023syncdreamer, stable-zero123}: Given a serial of reference images $I^{ref} = \{ I_i^{ref}  \}_1^\mathcal{N} $, reference poses $P^{ref} = \{ p_i^{ref} \}_1^\mathcal{N}$, target images $I^{target} = \{ I_i^{target}  \}_1^\mathcal{N} $, and target poses $P^{target} = \{ p_i^{target} \}_1^\mathcal{N}$, recent 3D generative models, such as Zero123~\citep{zero123}, Syncdreamer~\citep{liu2023syncdreamer}, and Stable-Zero123~\citep{stable-zero123}, take relative poses and reference images as inputs and generate target images. However, these models cannot generalize well to real car objects since they are trained on large-scale synthetic datasets~\citep{deitke2023objaverse, deitke2024objaverse}. In this work, we will demonstrate that our dataset can improve the robustness of these generative models to real cars.

\noindent\textbf{Single Image to 3D Generation}~\citep{dreamfusion,tang2023dreamgaussian}: Given a text prompt or single image, recent 3D generation methods generate 3D objects with Score Distillation Sampling (SDS)~\citep{dreamfusion} and diffusion generative models~\citep{zero123, stable-zero123}. However, these methods cannot generate high-quality 3D cars due to the lack of the prior of real cars in 3D-based diffusion models. Therefore, we would demonstrate the value of our dataset by improving recent 3D generation for real cars.

%% file: sec/exp.tex
\begin{figure*}[t] 

    \centering

    \includegraphics[width=0.85\textwidth]{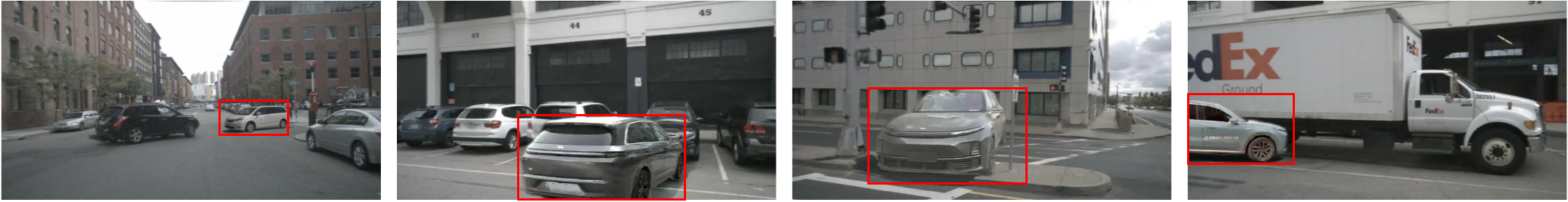} 
    \caption{\textbf{The simulated corner-case scenes.} 
    These scenes are rare but very important in real life.
    We use a red rectangle to highlight the simulated vehicles. These corner-case scenes show some vehicles have potential risks to traffic safety.
    }
    \label{simulate}
    \vspace{-5mm}
\end{figure*}

\begin{figure*}[!t] 

    \centering

    \includegraphics[width=0.85\textwidth]{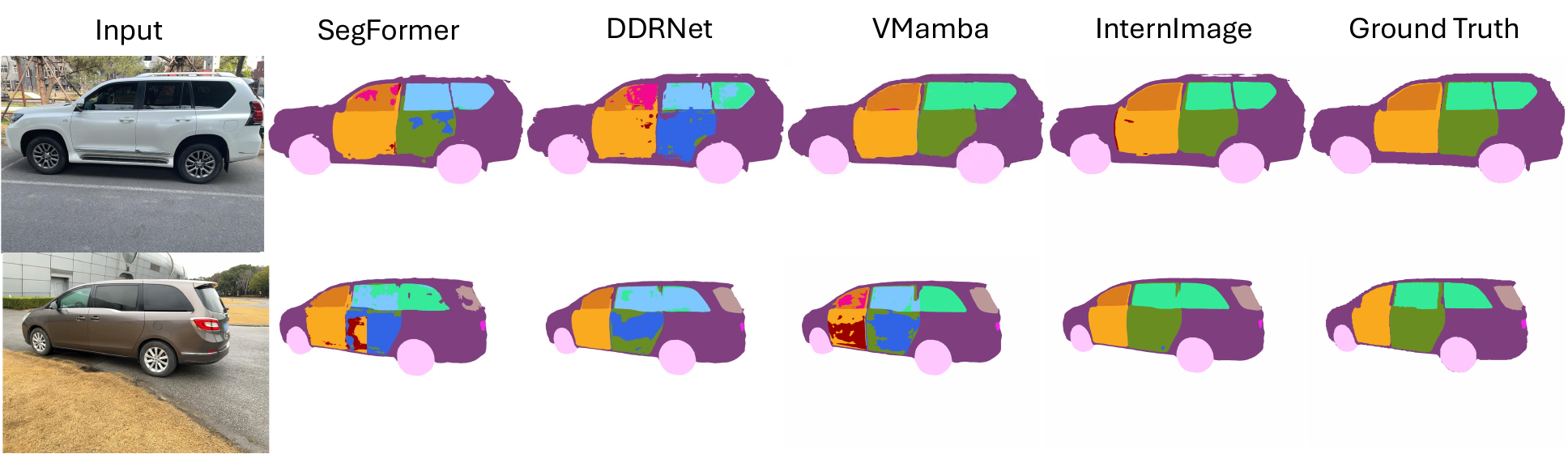} 
 \caption{\textbf{Qualitative comparisons among recent advanced image segmentation methods.} We select the inputs from the testing set of our images and evaluate the capacity of car component understanding for each method. }
    \label{seg_vis}
    \vspace{-5mm}
\end{figure*}

\section{Experiments}

\subsection{Experimental Setup}

\noindent\textbf{Corner-case 2D Detection. } In this task, we leverage the reconstructed cars to simulate rare and corner-case scenes. To be specific, we use Nuscenes~\citep{nuscenes} as background to simulate corner-case scenes with reconstructed cars and leverage recent popular detectors, like YOLOv8~\citep{yolov8}, as detectors for evaluation. To evaluate the robustness of detectors in corner-case scenes, we use the test part of the corner-case dataset, CODA~\citep{li2022coda} as a testing set. Since we focus on the corner-case scenes of cars, so we only evaluate a car class.

\noindent\textbf{2D Car Parsing.} In this task, we utilize  DDRNet~\citep{ddrnet}, SegFormer~\citep{xie2021segformer},  VMamba~\citep{liu2024vmamba}, and
InternImage~\citep{wang2023internimage}to benchmark our dataset. To be specific, we split 80\% of our car parsing maps in 3DRealCar as the training set and the rest of 20\% as the testing set.

\noindent\textbf{Neural Field-based Novel View Synthesis.}
In this task, we randomly choose 100 instances from each lighting condition in our dataset and split 80\% of the views per instance as the training set and the rest of 20\% as the testing set. Specifically, we employ recent state-of-the-art neural field methods, including Instant-NGP~\citep{instant-ngp}, 3DGS~\citep{kerbl2023gaussiansplatting}, GaussianShader~\citep{jiang2023gaussianshader}, and 2DGS~\citep{2dgs} to benchmark our dataset.

\noindent\textbf{Diffusion-based Novel View Synthesis.}
We finetune Zero123-XL~\citep{zero123} on our 3DRealCar dataset to enhance its generalization to real cars. Note that since the training of diffusion-based models needs entire objects centered on images, we use the images rendered by our trained 3D models as training images.

\noindent\textbf{Single Image to 3D Generation.}
In this task, we exploit Dreamcraft3D~\citep{sun2023dreamcraft3d} as our baseline. 
Dreamcraft3D exploits Stable-Zero123~\citep{stable-zero123} as a prior source for providing 3D generative prior. By fine-tuning Stable-Zero123 on our dataset,  we enable it to obtain car-specific prior so it generalizes well to real cars.

\subsection{Evaluation Metrics}

\noindent\textbf{PSNR $\uparrow$}: Peak Signal-to-Noise Ratio (PSNR) is a metric of the peak error between the original and a compressed or reconstructed image. Higher PSNR values indicate better image quality, with a higher similarity to original images. 

\noindent\textbf{SSIM  $\uparrow$}: Structural Similarity Index (SSIM)is a perceptual metric that considers changes in structural information, luminance, and contrast between the original and target image. Higher SSIM values indicate better performance.

\noindent\textbf{LPIPS $\downarrow$}: Learned Perceptual Image Patch Similarity (LPIPS) is a metric that uses deep learning models to assess the perceptual similarity between images. Lower LPIPS values indicate higher perceptual similarity. Unlike PSNR and SSIM, LPIPS leverages the capabilities of neural networks to better align with human visual perception.

\noindent\textbf{mAP  $\uparrow$}: In the object detection task, mAP denotes mean Average Precision, a widely used metric to evaluate the performance of detection algorithms. It measures the accuracy of the detector by considering both the precision and recall at different thresholds. Higher mAP means better results.

\subsection{2D Tasks}

\textbf{Corner-case 2D Detection.}
To obtain a reliable detector for corner-case scenes, we simulate corner-case scenes with reconstructed cars and synthesize them within a background. Specifically, we leverage a recent large-scale self-driving dataset, Nuscenes~\citep{nuscenes} to provide background information.
After obtaining a simulated corner-case dataset, we can use this dataset to train and obtain a reliable detector robust for corner-case scenes. 
As shown in Table \ref{t3}, we employ YOLOv5 and YOLOv8 serial models, CO-DETR~\cite{codetr}, and YOLOv12~\cite{tian2025yolov12} as our detectors for evaluation. To evaluate the performance of models in corner-case scenes, we leverage the test part of the CODA dataset~\citep{li2022coda} as our testing set.
In particular, when we increase the training simulated data from 500 to 5,000, the performance of detectors improves by a large margin. This phenomenon demonstrates that our simulated data is effective in improving detection performance to corner-case scenes. 
We provide the visualizations of simulated corner-case scenes in Figure \ref{simulate}. The detailed simulation process and more visualizations can be seen in the supplementary.

\input{table/detect}

\input{table/segment_3drec}

\noindent\textbf{2D Car Parsing.} We conduct benchmarks for car parsing maps of our dataset using recent segmentation methods, such as DDRNet\citep{ddrnet},  SegFormer\citep{xie2021segformer},
VMamba~\cite{liu2024vmamba}, and
InternImage~\cite{wang2023internimage}. The quantitative performance for these methods on our dataset is summarized in Table \ref{parsing_table}. Visual comparisons are provided in Figure \ref{seg_vis}. Our high-quality dataset enables these methods to achieve promising performance, highlighting its potential for application in self-driving systems. In particular, our car parsing annotations encourage self-driving systems to recognize different components of cars in practical scenarios for safer automatic decisions.
We believe that our detailed car parsing annotations could significantly contribute to advancing self-driving tasks.

\input{table/benchmark}
\input{table/3dgen}

\subsection{3D Tasks}

\textbf{Neural Field-based Novel View Synthesis.}
As depicted in Table \ref{t2}, we show  benchmark results of recent state-of-the-art neural field methods, such as Instant-NGP~\citep{instant-ngp}, 3DGS~\citep{kerbl2023gaussiansplatting}, GaussianShader~\citep{jiang2023gaussianshader}, 2DGS~\citep{2dgs}, Pixel-GS~\cite{pixel-gs}, and 3DGS-MCMC~\cite{mcmc} on our dataset. To the standard lighting condition, we can find that recent methods are capable of achieving PSNR more than 27 dB, which means these methods can reconstruct relatively high-quality 3D cars from our dataset. However, the reflective and dark condition results are lower than the standard. These two parts of our 3DRealCar bring two challenges to recent 3D methods. The first challenge is the reconstruction of specular highlights. Due to the particular property of cars, materials of car surfaces are generally glossy, which means it would produce plenty of specular highlights if cars are exposed to the sun or strong light. The second challenge is the reconstruction in a dark environment. The training images captured in the dark environment lose plenty of details for reconstruction. Therefore, how to achieve high-quality reconstruction results from these two extremely lighting conditions is a challenge to recent methods.  3D visualizations can be found on our project page.
We hope these results can encourage subsequent research for the 3D reconstruction in low-quality lighting conditions.

\begin{figure}[t] 
    \centering
    \includegraphics[width=0.48\textwidth]{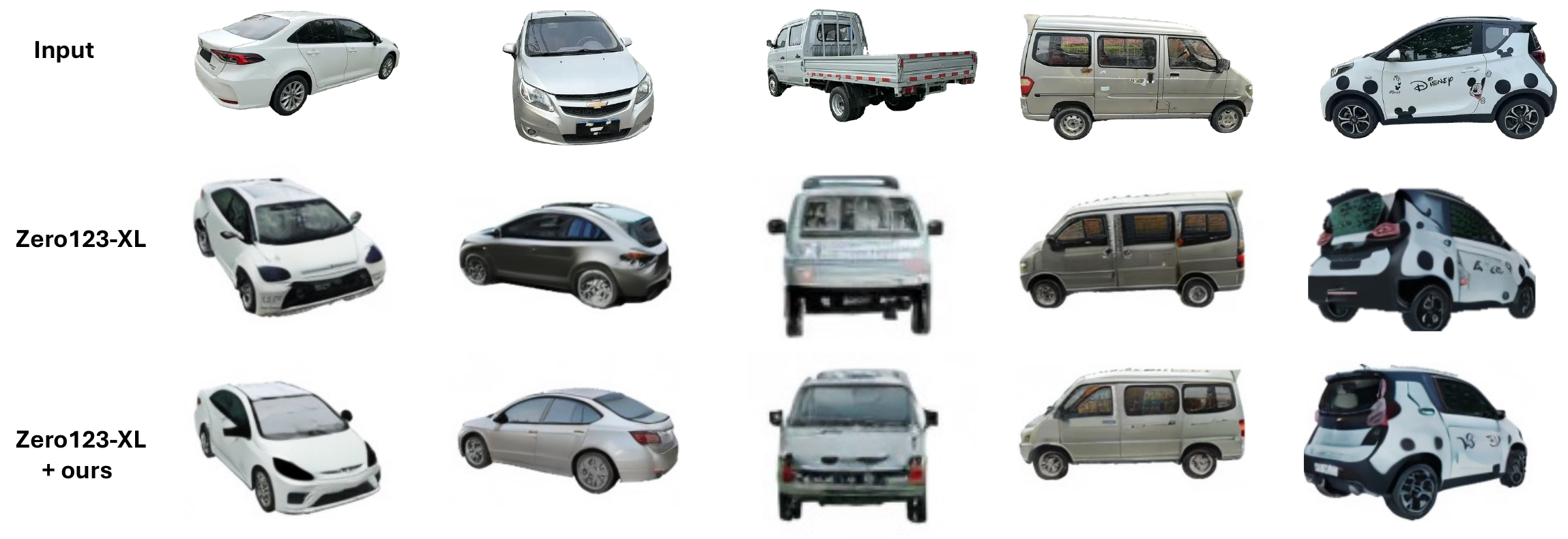} 
 \caption{\textbf{Visualizations of diffusion-based novel view synthesis. } we compare the results of the recent state-of-the-art diffusion-based method, Zero123-XL~\citep{zero123} and its improvement by training on our dataset. Our dataset provides car-specific prior for the generative model to generate more photorealistic car images. }
\label{vis_nsv}
\vspace{-5mm}
\end{figure}

\noindent\textbf{Diffusion-based Novel View Synthesis.}
As illustrated in Figure \ref{vis_nsv}, we show visual comparisons of Zero123-XL~\citep{zero123} and our improved version by training on our dataset. As we can see, given input images, we use Zero123-XL and our improved version to synthesize novel views. We can find that Zero123-XL prefers to generate unrealistic texture and geometry, due to the lack of prior for real objects.  In contrast, our improved version of Zero123-XL can generate photorealistic geometry and texture, which demonstrates the effectiveness of our dataset.

\begin{figure}[!t] 

    \centering

    \includegraphics[width=0.48\textwidth]{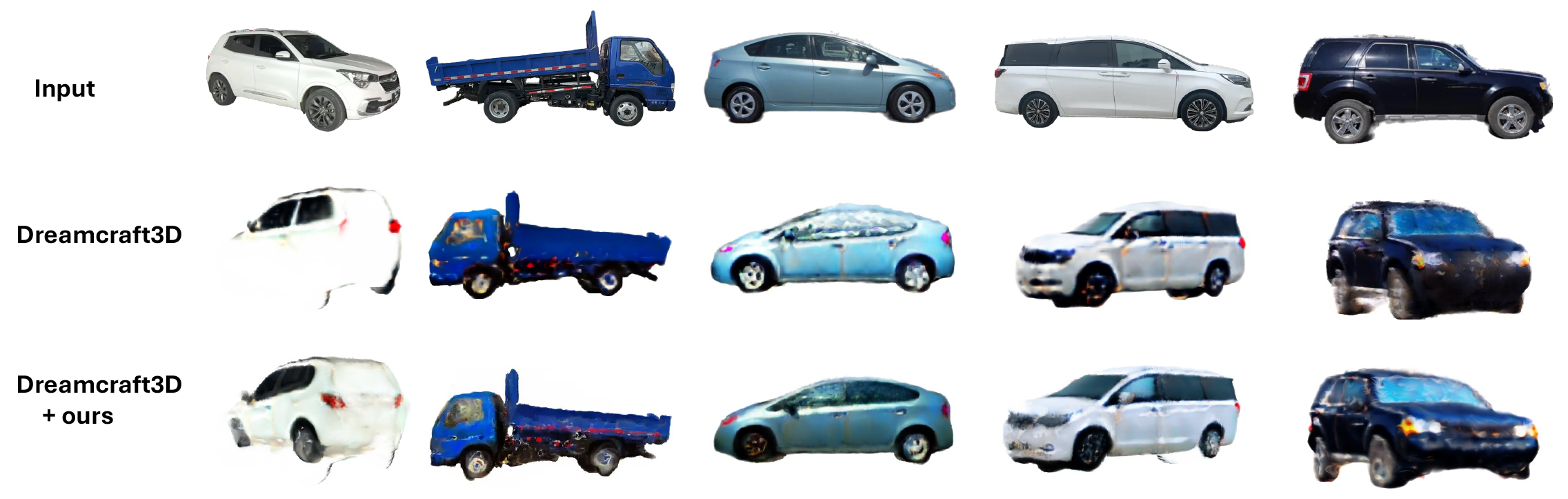} 
  \caption{\textbf{Visualizations of single-image-to-3D generation. } we compare the results of the recent state-of-the-art single-image-to-3D method, Dreamcraft3D~\citep{sun2024dreamcraft3d++} and is enhanced version by training on our dataset.  }
    \label{vis_3dgen}
        \vspace{-5mm}
\end{figure}

\noindent\textbf{Single Image to 3D Generation.}
Not only do we enhance novel view synthesis for diffusion-based models, but we also demonstrate improvements in 3D generation.
As depicted in Figure \ref{vis_3dgen}, we visualize 3D generation results of the recent state-of-the-art single-image-to-3D method, Dreamcraft3D~\citep{sun2024dreamcraft3d++}, along with its improved version by our dataset.  This figure shows that Dreamcraft3D sometimes fails to generate complete geometry or realistic texture, due to the scarcity of the real car prior. As shown in Table \ref{3dgen_table}, we also show quantitative comparisons of Dreamcraft3D and its improved version. CLIP-I means the similarity of rendered images with the original input. 
The quantitative and qualitative results indicate our dataset significantly improves 3D generation performance typically in terms of geometry and texture. These results underscore the effectiveness of our 3DRealCar dataset.


%% file: table/detect.tex
\begin{table}[t]

\caption{\textbf{ Detection improvements by simulated data for corner-case scenes.} We leverage lightweight YOLO serials models and recent state-of-the-art models for evaluation. We report the metric by calculating mAP@0.5 on the CODA dataset~\cite{li2022coda}.
}
\label{t3}
\centering


\resizebox{1\linewidth}{!}{
\begin{tabular}{c|cccccc}
\toprule \toprule
Simulated Data & YOLOv5n        & YOLOv5s        & YOLOv8n        & YOLOv8s        & CO-DETR           &YOLOv12x \\ \midrule 
1000           & 0.285          & 0.341          & 0.299          & 0.371          & 0.465           &0.412\\
2000           & 0.304          & 0.357          & 0.312          & 0.366          & 0.481           &0.441\\
3000           & 0.345          & 0.389          & 0.357          & 0.403          & 0.517           &0.489\\
4000           & 0.357          & 0.408          & 0.386          & 0.413          & 0.551           &0.531\\
5000           & \textbf{0.361} & \textbf{0.426} & \textbf{0.386} & \textbf{0.435} & \textbf{0.582}  &\textbf{0.565}\\ \bottomrule
\end{tabular}}
    \vspace{-3mm}

\end{table}

%% file: table/segment_3drec.tex
\begin{table}[t]
  \centering

    \caption{\textbf{Benchmark results on 2D car parsing of our 3DRealCar dataset.} We use recent advanced image segmentation methods~\cite{wang2023internimage, liu2024vmamba,ddrnet, xie2021segformer} to benchmark our dataset.}
  \centering
  \vspace{-3mm}
    \resizebox{0.9\linewidth}{!}{
    \begin{tabular}{lcccc}
        \toprule \toprule
        \textbf{Method}  &SegFormer&  DDRNet  & VMamba  & InternImage\\ \midrule
\textbf{mIOU}   $\uparrow$    &     0.541     &  0.606    & 0.610 & \textbf{0.671}     \\ \hline
\textbf{mAcc}   $\uparrow$      &    0.652      &  0.732     &   0.734   &  \textbf{0.786} \\ \bottomrule
        \end{tabular}
    }
    \label{parsing_table}

    \vspace{-3mm}
\end{table}

%% file: table/benchmark.tex
\begin{table*}[t]

\caption{\textbf{Benchmark results on 3D reconstruction of our 3DRealCar dataset.} We present the 3D reconstruction performance of recent state-of-the-art methods in three lighting conditions, standard, reflective, and dark, respectively.
The best results are highlighted.
}
\vspace{-3mm}
\label{t2}
\centering

\resizebox{0.95\linewidth}{!}
{
\begin{tabular}{cccccccccc}
\toprule \toprule
\multirow{2}{*}{\textbf{Method}} & \multicolumn{3}{c}{\textbf{Standard}}              & \multicolumn{3}{c}{\textbf{Reflective}}            & \multicolumn{3}{c}{\textbf{Dark}}                  \\
                                 & \textbf{PSNR $\uparrow$}  & \textbf{SSIM$\uparrow$}   & \textbf{LPIPS$\downarrow$}  & \textbf{PSNR$\uparrow$}  & \textbf{SSIM$\uparrow$}   & \textbf{LPIPS$\downarrow$}  & \textbf{PSNR$\uparrow$}  & \textbf{SSIM$\uparrow$}   & \textbf{LPIPS$\downarrow$}  \\ \midrule
Instant-NGP~\cite{instant-ngp}                      & 27.31          & 0.9315          & 0.1264          & 24.37          & 0.8613          & 0.1962 & 23.17          & 0.9152          & 0.1642          \\
3DGS~\cite{kerbl2023gaussiansplatting}                             & 27.47          & 0.9367 & 0.1001 & 24.58          & 0.8647          & 0.1852 & 23.51 & 0.9181 & 0.1613 \\
GaussianShader~\cite{jiang2023gaussianshader}                   & 27.53         & 0.9311          & 0.1109          & \textbf{25.41} & \textbf{0.8684} & \textbf{0.1423} & 23.39          & 0.9172          & 0.1631          \\
2DGS~\cite{2dgs}                             & 27.34 & 0.9341          & 0.1095          & 23.19          & 0.8509          & 0.2041 & 22.63          & 0.9148          & 0.1681          \\
Pixel-GS~\cite{pixel-gs}                             & \textbf{27.67} & \textbf{0.9391}         & 0.0994          & 24.81          & 0.8659          & 0.1541 & 23.54          & 0.9174          & \textbf{0.1617}          \\
3DGS-MCMC~\cite{mcmc}                             & 27.63 & 0.9382          & \textbf{0.0986}          & 24.92         & 0.8681          & 0.1621 & \textbf{23.63}          & \textbf{0.9198 }         & 0.1622          \\

\bottomrule

\end{tabular}
}

\vspace{-3mm}

\end{table*}

%% file: table/3dgen.tex
\begin{table}[t]

   \caption{\textbf{Quantitative comparisons of SOTA 3D Generation method, Dreamcraft3D~\cite{sun2023dreamcraft3d} and its improved version by trained on our dataset. } CD denotes Chamfer Distance.  }
   \vspace{-3mm}
    \centering
    \scriptsize
    \def\arraystretch{1.2}  
     \resizebox{0.85\linewidth}{!}{
        \begin{tabular}{l|ccc}
        \toprule \toprule
          \textbf{Method}   & \textbf{CLIP-I} $\uparrow$ & \textbf{Hausdorff} $\downarrow$  & \textbf{CD} $\downarrow$ \\ \midrule
        Dreamcraft3D~ &      0.812        &   1.572  &        0.587    \\ \hline
        +our dataset        &      \textbf{0.847}       &   \textbf{1.364}   & \textbf{0.371}          \\ \bottomrule
        \end{tabular} 
        }
    \label{3dgen_table}
  \vspace{-3mm}
\end{table}

%% file: sec/conclusion.tex
\section{Conclusion}

In this paper, we propose the first large-scale high-quality 3D real car dataset, named 3DRealCar. 
The collected dense and high-resolution 360-degree views for each car can be used to reconstruct a high-quality 3D car. 
Extensive experiments demonstrate the efficacy and challenges of our 3DRealCar in 3D reconstruction. Thanks to the reconstructed high-quality 3D cars from our dataset and car-part level annotations, our dataset can be utilized to support various tasks related to cars. 
In addition, the benchmarking results can serve as baselines for prospective research.
Although 3DRealCar currently only has car exterior views, we intend to provide both exterior and interior views in the future to further promote the reconstruction of more intact 3D cars.